\documentclass[letterpaper, 10 pt, conference]{ieeeconf}  % Comment this line out if you need a4paper

\IEEEoverridecommandlockouts                              % This command is only needed if 
\usepackage{hyperref}
\usepackage{cite}
\usepackage{graphicx}
\usepackage{float}
\usepackage{amsfonts}
\usepackage{amsmath}
\usepackage{algorithm}
\usepackage{algpseudocode}
\usepackage{tabularx}
\usepackage{multirow}
\usepackage{booktabs}

\title{\LARGE \bf
Beyond \textit{Bare} Queries:\\ Open-Vocabulary Object Grounding with 3D Scene Graph
}

\author{Sergey Linok$^{1,*,\dag}$, Tatiana Zemskova$^{1,2,*}$, Svetlana Ladanova$^{1}$, Roman Titkov$^{1}$, Dmitry Yudin$^{1,2}$, \\ Maxim Monastyrny$^{3}$, Aleksei Valenkov$^{3}$% <-this % stops a space
\thanks{$^{1}$Center for Cognitive Modeling, Moscow Institute of Physics and Technology, Dolgoprudny, Russia}% <-this % stops a space
\thanks{$^{2}$AIRI, Moscow, Russia}% <-this % stops a space
\thanks{$^{3}$Sberbank of Russia, Robotics Center, Moscow, Russia}% <-this % stops a space
\thanks{$^{\dag}$Corresponding author: linok.sa@phystech.edu}
\thanks{$^{*}$Equal contribution}
}

\begin{document}

\maketitle
\thispagestyle{empty}
\pagestyle{empty}

%%%%%%%%%%%%%%%%%%%%%%%%%%%%%%%%%%%%%%%%%%%%%%%%%%%%%%%%%%%%%%%%%%%%%%%%%%%%%%%%
\begin{abstract}

Locating objects described in natural language presents a significant challenge for autonomous agents. Existing CLIP-based open-vocabulary methods successfully perform 3D object grounding with simple (\textit{bare}) queries, but cannot cope with ambiguous descriptions that demand an understanding of object relations. To tackle this problem, we propose a modular approach called BBQ (Beyond \textit{Bare} Queries), which constructs 3D scene graph representation with metric and semantic spatial edges and utilizes a large language model as a human-to-agent interface through our deductive scene reasoning algorithm. BBQ employs robust DINO-powered associations to construct 3D object-centric map and an advanced raycasting algorithm with a 2D vision-language model to describe them as graph nodes. On the Replica and ScanNet datasets, we have demonstrated that BBQ takes a leading place in open-vocabulary 3D semantic segmentation compared to other zero-shot methods. Also, we show that leveraging spatial relations is especially effective for scenes containing multiple entities of the same semantic class. On challenging Sr3D+, Nr3D and ScanRefer benchmarks, our deductive approach demonstrates a significant improvement, enabling objects grounding by complex queries compared to other state-of-the-art methods. The combination of our design choices and software implementation has resulted in significant data processing speed in experiments on the robot on-board computer. This promising performance enables the application of our approach in intelligent robotics projects. We made the code publicly available at \href{https://linukc.github.io/BeyondBareQueries/}{linukc.github.io/BeyondBareQueries}.

\end{abstract}

%%%%%%%%%%%%%%%%%%%%%%%%%%%%%%%%%%%%%%%%%%%%%%%%%%%%%%%%%%%%%%%%%%%%%%%%%%%%%%%%
\section{INTRODUCTION}

Open-vocabulary 3D perception is a primary challenge for advanced AI-powered autonomous agents. For instance, locating a referred object based on a complex text query in an environment full of semantically similar distractors remains an unsolved question. 

The effective combination of vision and text modalities has been a subject of intensive study in the research field, with a particular focus on CLIP-based \cite{radford2021learning,mu2022slip,fang2023eva,sun2024eva,zhai2023sigmoid,sun2023alpha} encoders and more advanced visual foundation models \cite{kirillov2023segment,zou2024segment,liu2023llava,girdhar2023imagebind,li2023blip,ranzinger2023radio,wu2023incremental,tsimpoukelli2021multimodal,li2023blip,liu2023llava,liu2023improvedllava,liu2024llavanext}. To address object inter-relations in complex 3D environments, scene graph representations have emerged as the most promising approach \cite{armeni20193d,koch2024open3dsg,werby2024hierarchical,gu2023conceptgraphs}. Scene edges are described as embeddings or text, allowing rich and fine-grained representations of spatial and semantic relationships. Large language models (LLMs) are increasingly being utilized in 3D perception as tools capable of sophisticated reasoning \cite{gu2023conceptgraphs,chen2022leveraging,werby2024hierarchical,kim2023llm4sgg,hong20233d,huang2023chat,fu2024scene,huang2024language}. Leveraging the power of scene graph representation as input, such systems can robustly interpret queries with the power of inner knowledge and accurately respond about scene-specific objects or locations, opening up new possibilities for applications in robotics, augmented reality, and intelligent assistive technologies. Despite the impressive qualities exhibited by these approaches, significant challenges remain in the field.

\begin{figure}[t]
  \centering
  \includegraphics[width=1.0\linewidth]{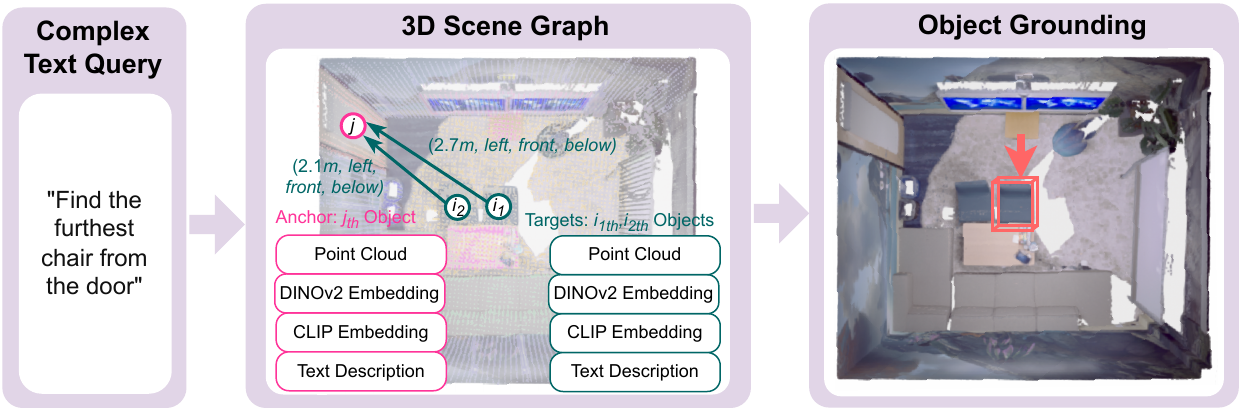}
  \vspace{-3mm}
  \caption{\label{ga}Proposed \textit{BBQ} approach leverages foundation models for high-performance construction of an object-centric class-agnostic 3D map of a static indoor environment from a sequence of RGB-D frames with known camera poses and calibration. To perform scene understanding, we represent environment as a set of nodes with spatial relations. Utilizing a designed \textit{deductive} scene reasoning algorithm, our method enable efficient natural language interaction with a scene-aware large language model.}
\end{figure}

In our research, we particularly focus on approaches that exploit generalization ability of pretrained models to perform open-vocabulary 3D grounding in a zero-shot manner from an RGB-D sequence. The first challenge for the aforementioned methods is the difficulty in accumulating reliable visual/text representations for scene objects with a low resource cost on a robot on-board computers. The second challenge is to preserve scene spatial awareness in both view-dependent and view-independent object relations. The third challenge is how to effectively incorporate 3D scene representation and query a large language model to perform the desired task.

To address these challenges, we explicitly separate the 3D objects mapping process from object visual/text encoding. Self-supervised DINO image embeddings have proven to be robust discriminative features that are sufficient for strong 2.5D understanding \cite{banani2024probing,tumanyan2024dino,keetha2023anyloc,amir2021deep}. For each frame, we extract these deep features and employ fast DINO-based proposals-to-objects accumulation (Sec. \ref{objects_map}). At the end of a sequence, to select a projection frame for 2D captioning we perform a designed multiview clustering technique (Sec. \ref{3Dto2D}), that drastically reduces the search space (Sec. \ref{nodes}).

In order to maintain spatial awareness in a scene graph, we incorporate both metric and semantic spatial edges (Sec. \ref{edges}). Inspired by the chain-of-thought technique \cite{wei2022chain}, we propose a two-stage deductive scene reasoning algorithm for object grounding (Sec. \ref{how_to_apply}). In the first stage, an LLM selects potential target and anchor objects for a user query based solely on object descriptions (Sec. \ref{how_to_apply}, Fig.~\ref{ga}).
This not only reduces the search space but also minimizes the number of edges constructed. In the second stage, the LLM makes the final prediction, utilizing additional information regarding the spatial positions of both targets and anchors, as well as the edges connecting them.

\textbf{To summarize}, the contributions of this paper include:
\begin{itemize}
  \item high-performance DINO-based 3D object-centric map construction algorithm from a sequence of posed \mbox{RGB-D} images (Sec. \ref{objects_map}, Sec. \ref{3Dto2D});
  \item cheap to construct and descriptive rich 3D scene graph representation with metric and semantic spatial edges (Sec. \ref{nodes}, Sec. \ref{edges});
  \item \textit{deductive} scene reasoning algorithm to use our scene graph representation with large language model for open-vocabulary 3D object grounding (Sec. \ref{how_to_apply});
  \item integration of these capabilities in a single modular method called BBQ. We make the code publicly available at \href{https://linukc.github.io/BeyondBareQueries/}{linukc.github.io/BeyondBareQueries}.
\end{itemize}

\section{RELATED WORKS}

\subsection{Open-Vocabulary 3D Segmentation}

Open-vocabulary segmentation gives the ability to locate an arbitrary semantic category on a scene. The key challenge in working with 3D data is efficient encoding of aligned visual/text embeddings \cite{radford2021learning,mu2022slip,fang2023eva,sun2024eva,zhai2023sigmoid,sun2023alpha,kirillov2023segment,zou2024segment,liu2023llava,girdhar2023imagebind,li2023blip,ranzinger2023radio,wu2023incremental} in scene reconstruction. Most modern methods project 2D features into 3D \cite{tsagkas2023vl,ha2022semantic,conceptfusion,lu2023ovir,gu2023conceptgraphs,yamazaki2023open,hong20233d}, perform 2D-to-3D pointwise distillation \cite{huang2023openins3d,huang2023chat}, combine both approaches \cite{nguyen2023open3dis,yang2023regionplc,takmaz2023openmask3d,peng2023openscene,koch2024open3dsg,fu2024scene,yin2023sai3d,werby2024hierarchical}, or use advanced scene-specific representations such as NeRF \cite{kerr2023lerf,liu2023weakly,liao2024ov,tie2024o2v,ying2023omniseg3d} and Gaussian Splatting \cite{shi2023language,qin2023langsplat,jiang2023feature} with auxiliary 2D supervision. To obtain reliable 2D features that describe more than just the local context, various techniques are applied, such as sliding window averaging~\cite{takmaz2023openmask3d}, class-agnostic mask cropping~\cite{gu2023conceptgraphs}, superpixel grouping~\cite{guo2023sam}. 

Final 3D features can be represented in different format: one feature by point~\cite{conceptfusion}, voxel~\cite{zhu2024open}, object~\cite{gu2023conceptgraphs}. The last format allows the answer to be aligned with some scene instance with the closest similarity to the query. In contrast, other representations will always return some part of the environment with a similarity higher than a defined threshold.

\subsection{3D Scene Graphs}

3D scene graph provides a compact and structured representation of the environment, capturing not only the objects  but also their spatial arrangements and semantic relationships. 3D scene graph can be obtained from known 3D geometries \cite{werby2024hierarchical,armeni20193d,hughes2022hydra} or fused from RGB sequence as a combination of local 2D scene graphs \cite{kim20193,wu2023incremental}.
Nodes and edges can belong to a fixed set of semantic categories~\cite{wang2023vl, zhang2024egosg} or be predicted in an open-vocabulary manner~\cite{koch2024open3dsg, chen2024clip, gu2023conceptgraphs} allowing transfer to unseen environments without additional fine-tuning.
Graphs can describe one-level relations (for example, objects-objects) \cite{kim20193,wald2020learning,wald2022learning} or contain a hierarchical representation \cite{werby2024hierarchical,honerkamp2024language,hughes2022hydra}.

\subsection{3D Object Grounding}

3D object grounding aims to find a particular instance in a scene, described in a natural language query. The query can contain references to other objects or a specific location~\cite{chen2020scanrefer, achlioptas2020referit_3d}, an affordance reference~\cite{majumdar2024openeqa, zhu2024empowering}, or other types of object attributes~\cite{achlioptas2020referit_3d}.
Solving this task involves understanding the relationships between objects. It can be done with interpretable 3D scene graphs \cite{li2024r2g,gu2023conceptgraphs} or learned with supervision from annotation \cite{zhang2023multi3drefer,zhao20213dvg,hsu2023ns3d}.

Integrating an LLM into a perception pipeline can provide a source of domain knowledge that enables complex query understanding and reasoning. Different approaches use an LLM simply as an interface, providing scene-specific description in an input context \cite{yuan2024visual,yang2024llm}, or performing fine-tuning \cite{chen2022language,zhu2024empowering, chen2024grounded, huang2023chat}.
To add the ability to perceive the visual modality, these methods use pre-trained point cloud encoders \cite{chen2022language, zhu2024empowering, huang2023chat} or train them to align with the text modality~\cite{chen2024grounded}. Then the authors perform instruction tuning of the projection layers~\cite{chen2022language, huang2023chat, chen2024grounded}  or additionally fine-tune the LLM~\cite{zhu2024empowering}.

Unlike existing 3D Object Grounding methods that integrate LLMs, our approach does not rely on ground-truth scene point clouds or domain-specific fine-tuning~\cite{miyanishi2024cross3dvg}. Instead, we build a 3D scene graph on-the-fly from a sequence of posed RGBD images, using only foundation models, enabling easy transfer to unseen environments. We also show that our method, combined with the \textit{deductive} scene reasoning algorithm outperforms the similar baselines ConceptGraphs~\cite{gu2023conceptgraphs} and LLM-Grounder~\cite{yang2024llm}.

Our work is conceptually closest to ConceptGraphs~\cite{gu2023conceptgraphs}, however each of our contribution identifies gap in existing methodology and suggest novel solution. We show in the experiment section that it is not only improve overall quality (Sec.~\ref{experiments}), but also significantly reduce required computational resources and allow to apply BBQ in real-life scenarios (Sec.~\ref{real_robot}).

\begin{figure*}[t]
\includegraphics[width=0.97\textwidth]{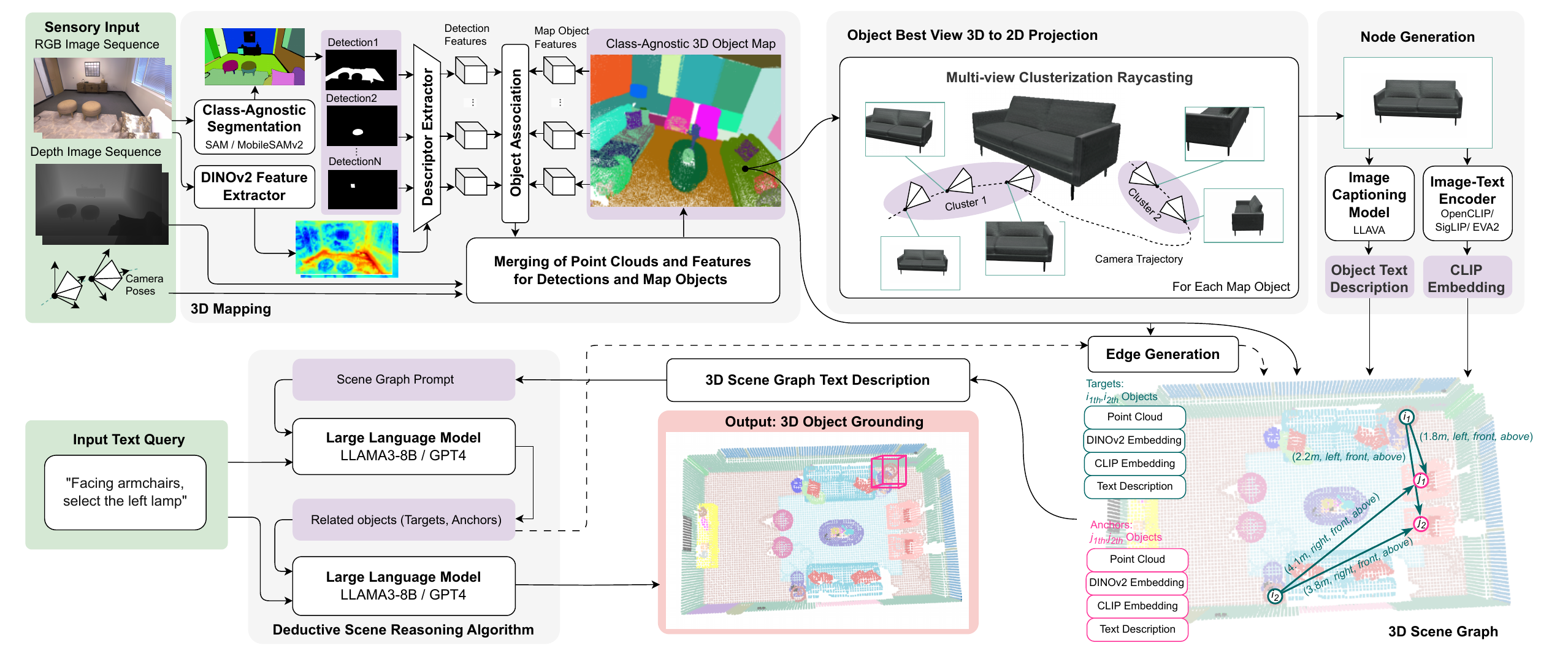}
\centering
\caption{\label{method_fig} An object-centric class-agnostic 3D map is iteratively constructed from a sequence of RGB-D camera frames and their poses by associating 2D MobileSAMv2 mask proposals with 3D objects with deep DINOv2 visual features and spatial constraints (Sec. \ref{objects_map}). To visually represent objects after building the map, we select the \textit{best} view based on the largest projected mask from $L$ cluster centroids that represent areas of object observations (Sec. \ref{3Dto2D}). We leverage LLaVA1.6 \cite{liu2023improvedllava} and text-aligned visual encoder EVA2~\cite{fang2023eva} to describe object visual properties (Sec. \ref{nodes}). With the node's text descriptions, spatial locations, metric and semantic spatial edges (Sec. \ref{edges}) we utilize LLM in our \textit{deductive} reasoning algorithm (Sec. \ref{how_to_apply}) to perform a 3D object grounding task.}
\end{figure*}

\section{METHOD}

\subsection{Object-centric 3D map construction (3D Mapping)}
\label{objects_map}

For each input RGB-D frame $I$ with a resolution $H \times W$ pretrained foundation model extract set of 2D proposal (boolean masks) $M \in \mathbb{R}^{H \times W}$ and DINO embeddings $E_{DINO} \in \mathbb{R}^{H/s \times W/s \times dim}$, where $s$ is DINO transformer patch $stride$. For $M$ we experimented with MobileSAMv2 \cite{zhang2023mobilesamv2}, for $E_{DINO}$, we examined DINOv2 with registers \cite{darcet2023vision}.

After a series of filtration checks to discard low confidence, small or too large regions, each passed mask $m$ is represented as point cloud $p \in \mathbb{R}^{N \times 3}$, where $N$ - number of points in mask that we project based of depth information, pose and camera calibration. To point cloud $p$ we additionally apply DBSCAN to remove noise from inaccurate 2D proposal $m$. Associated descriptor $d \in \mathbb{R}^{dim}$ is extracted from $E_{DINO}$ by averaging all features in interpolated to ${R}^{H/s \times W/s}$ mask $m$. Each frame $I$ from sequence of images is represented by detection set $\lbrace(p_{k}, d_{k}) \mid k \in (1, \dots , K)\rbrace$, where $K$ - number of selected proposals.

To construct a class-agnostic object-centric 3D map on each frame we perform an association process between incoming detections set and objects on the scene. For the first frame, we simply initialize objects as detections. Association process consists of finding visual cosine similarities between all intersected instances. If for $j\text{-}th$ detection $CosineSimularity(d_{j}, d_{i})$ lower than visual threshold $\sigma_{vis}$ for every object $i$, we initialize this detection as a new object, else merge with closest by cosine similarity. During merging process we combine point clouds, increase number of detections for object, add frame index for future 3D to 2D projection search and update object visual descriptor with moving average. Assigning a higher weight to incoming descriptors leverages the capability of DINO features to effectively identify correspondences between local frames while preserving information for objects merging. To reduce the growing number of objects that appear after unsuccessful association each $m\text{-}th$, object merging with lower $\sigma_{vis}$ and spatial overlap is called. After object map construction we perform the postprocessing step: filter objects to discard outliers (low points, size or number of detections).

\subsection{Object \textit{best} view 3D to 2D projection}
\label{3Dto2D}

Because performing the raycasting procedure for all poses is time-consuming, we decided to cluster 3D camera coordinates $P_i = \{(x, y, z)_{q} \mid q \in (1, \dots , Q)\}$ from $Q$ viewpoints where we observed the object $o_i$, to $L$ groups. Then, we select $L$ poses closest to the centroids of the corresponding clusters, which represent centers of object observation. Our experiments demonstrate $L=5$ to be a sufficient number of observations on the indoor Replica and ScanNet datasets. Then we select \textit{best} object view based on the largest area of projection (Eq. \ref{eq:3Dto2D}):
\begin{equation}
\label{eq:3Dto2D}
\max_{P_{l} | l \in (1, \dots, L)} A\left(R(o_i, P_{l}) \right),
\end{equation}
where $R$ is the raycasting operator that converts an object's point cloud $p_i$ in the pose $P_{l}$ into mask, and $A$ is the operator that estimates the area of the object's mask.

\subsection{3D scene graph node generation}
\label{nodes}

To describe each object (node) $o_i$ we utilize a visual language model and text-aligned visual encoder on a projected crop. Our internal experiments show that LLaVA1.6~\cite{liu2023improvedllava} performs best for indoor scene captioning and EVA2~\cite{fang2023eva} for visual encoding. We call the captioning model with a prompt \textit{``Describe visible object in front of you, paying close attention to its spatial dimensions and visual attributes"}. We also incorporate enhanced environmental awareness in the prompt and several handcrafted examples.

\subsection{3D Scene graph edge generation}
\label{edges}

When generating scene graph edges, we pursue two goals: 1) enabling LLM to answer complex user queries, and 2) compactly describing object relations in textual form.

To achieve these goals, we define scene graph edges using two types of relations: metric relations $d_{ij}$ and semantic relations $s_{ij}$. Metric relations are defined as Euclidean distances between the centers of objects' bounding boxes. Semantic relations include the following set: \textit{left}, \textit{right}, \textit{front}, \textit{behind}, \textit{above} and \textit{below}. Since the relations left, right, front, behind are view-dependent, they are defined independently for each grounding phrase using the heuristic algorithm from ZSVG3D~\cite{yuan2024visual}. For each grounding phrase, we define target and anchor objects (see Sec.~\ref{how_to_apply} for more details). For each target-anchor pair, a virtual camera is placed at the room's center and pointed at the anchor. View-dependent relations are then determined based on the 2D projections of objects in the camera's egocentric view. Since the pairs of relations left/right, front/back, above/below are not mutually exclusive, a semantic spatial edge $s_{ij}$ consists of a set of three relations that include one relation from each pair (e.g. $s_{ij}=(left, front, above)$).

\begin{algorithm}[t]
\tiny
\caption{\textit{Deductive} scene reasoning algorithm $(O^{id}, O^{caption}, O^{center}, O^{extent}, query)$}
\label{llm_inf_alg}
\begin{algorithmic}
\State $(TargetIds, AnchorIds)  \gets LLM(query, O^{id}, O^{caption})$
\State $N_{target} \gets$ number of $TargetIds$
\State $N_{anchor} \gets$ number of $AnchorIds$

\State $RelatedObjects \gets \emptyset$
\For{$i$ in range $(0,N_{target})$}
\State $o^{relations}_i \gets \emptyset$
\For{$j$ in range $(0,N_{anchor})$}

$s_{ij} \gets  semantic\_relation(o^{center}_i, o^{center}_j)$

$d_{ij} \gets  euclidean\_distance(o^{center}_i, o^{center}_j)$

$E_{ij} \gets  (d_{ij}, s_{ij})$

Append $E_{ij}$ to $o^{relations}_i$

\EndFor

$Target \gets \lbrace o^{id}_i, o^{caption}_i, o^{center}_i, o^{extent}_i, o^{relations}_i\rbrace$

Append $Target$ to $RelatedObjects$
\EndFor

\For{$j$ in range $(0,N_{anchor})$}

$Anchor \gets \lbrace o^{id}_i, o^{caption}_i, o^{center}_i, o^{extent}_i\rbrace$

Append $Anchor$ to $RelatedObjects$

\EndFor

\State $FinalObjectId \gets LLM(query, RelatedObjects)$

\end{algorithmic}
\end{algorithm}

\subsection{Applying scene graph to a LLM}
\label{how_to_apply}

We propose using an LLM and scene text description to locate objects by their open vocabulary references. We store the scene description as a JSON file containing an objects list $O$. For each object $i$, we store its id ($o^{id}_i$), caption ($o^{caption}_i$), center ($o^{center}_i$) and extent ($o^{extent}_i$) of its 3D bounding box. We construct list $E$ of the objects connections for each user query independently. The relation between objects $o_i$ and $o_j$ includes $o^{caption}_i$, $o^{id}_i$,$o^{caption}_j$, $o^{id}_j$, the Euclidean distance between their bounding box centers ($d_{ij}$) and semantic relation ($s_{ij}$).

Real scenes may contain many objects (e.g., over 100), potentially forming a complete graph and resulting in a long scene description (over 32k symbols). This can degrade LLM performance due to the long context. However, retrieving a referred object typically doesn't require the location of all objects. Therefore, we propose a \textit{deductive} scene reasoning algorithm to find objects efficiently.

The \textit{deductive} scene reasoning algorithm involves several consecutive LLM calls. First, the LLM uses as input a scene description with object IDs $O^{id}$ and their caption $O^{caption}$, along with the user's query $query$. It selects target and anchor objects IDs ($TargetIds$ and $AnchorIds$) for answering the question in \textit{general} case based on the objects semantics. The target object is the object that answers the user's query. The anchor object is the object with which the target object has spatial relations in the user's question (see example in Fig.~\ref{method_fig}).
We create a scene description with additional location and connection details, but only for the selected $Targets$ and $Anchors$. This results in a compact description focused on the objects relevant to the user's query in the \textit{special} scene.

For each object, we include a ``relations" field in its description, listing sentences in the following template: ``The $o^{caption}_i$ with id $o^{id}_i$ is $s_{ij}$ and at distance $d_{ij}$ m  from the $o^{caption}_j$ with id $o^{id}_i$".
To limit scene description length, we add edge $e_{ij}$ to the scene graph only if $o_i \in TargetObjects$ and $o_j \in AnchorObjects$. Using the compact scene description of relevant objects, the LLM retrieves the final objects in JSON format. If the response needs formatting, we make another LLM call to adjust it.

Algorithm~\ref{llm_inf_alg} details our scene \textit{deductive} inference algorithm. Results in Sec.~\ref{scene_graph_results} demonstrate the benefits of incorporating spatial metric and semantic relations in the scene description and employing the \textit{deductive} scene reasoning algorithm for object grounding using LLM.

\begin{figure}[t]
  \centering
  \includegraphics[width=0.5\textwidth]{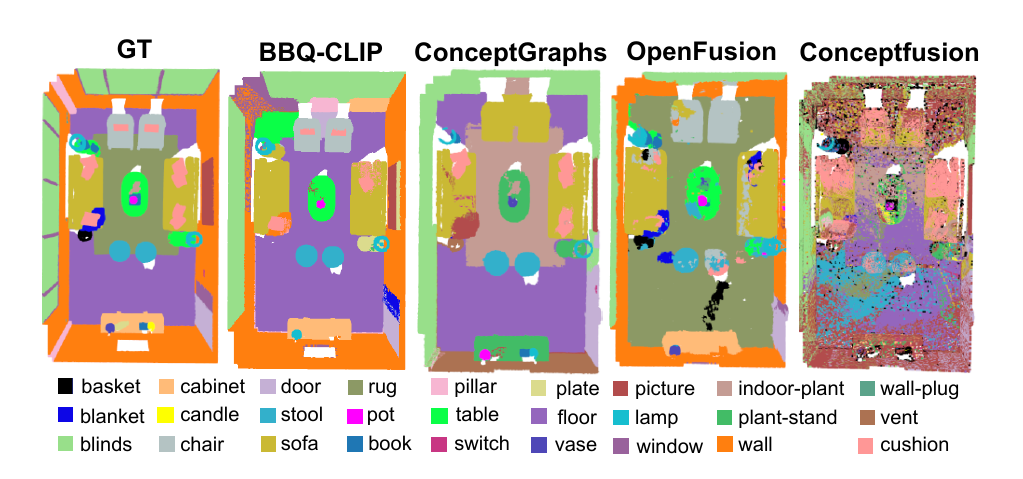}
  \vspace{-5mm}
  \caption{\label{replica_semseg_viz} Qualitative examples of 3D open-vocabulary semantic segmentation on the Replica.}
\end{figure}

\begin{table}[t]
  \scriptsize
  \caption{
    3D open-vocabulary semantic segmentation benchmark.
  }
  \label{tab:3Dsemseg}
  \centering
  \begin{tabularx}{0.45\textwidth}{lcXXXXXX}
    \toprule 
    \multirow{2}{*}{} &
      \multicolumn{1}{c}{} &
      \multicolumn{3}{c}{Replica} &
      \multicolumn{3}{c}{ScanNet} \\
    & Methods & mAcc\textuparrow & mIoU\textuparrow & fmIoU\textuparrow 
    & mAcc\textuparrow & mIoU\textuparrow & fmIoU\textuparrow \\
    \midrule
    \textit{\rotatebox[origin=c]{90}{\textit{Privileged}}} & OpenFusion  & 0.41 & 0.30 & 0.58 & 0.67 & 0.53 & 0.64 \\
    \midrule
    \multirow{4}{*}{\rotatebox[origin=c]{90}{\textit{Zero-Shot}}} & ConceptFusion  & 0.29 & 0.11 & 0.14 & 0.49 & 0.26 & 0.31\\
    & OpenMask3D  & - & - & - & 0.34 & 0.18 & 0.20\\
    & ConceptGraphs  & 0.36 & 0.18 & 0.15 & 0.52 & 0.26 & 0.29\\
    & BBQ-CLIP & \textbf{0.38} & \textbf{0.27} & \textbf{0.48} & \textbf{0.56} & \textbf{0.34} & \textbf{0.36}\\
  \bottomrule
  \end{tabularx}
\end{table}

\section{EXPERIMENTS}
\label{experiments}
We conduct our experiments on Replica\cite{straub2019replica} and ScanNet\cite{dai2017scannet} RGB-D data. For each datasets we select 8 scenes: (\textit{room0}, \textit{room1}, \textit{room2}, \textit{office0}, \textit{office1}, \textit{office2}, \textit{office3}, \textit{office4}) and (\textit{0011\_00}, \textit{0030\_00}, \textit{0046\_00}, \textit{0086\_00}, \textit{0222\_00}, \textit{0378\_00}, \textit{0389\_00}, \textit{0435\_00}) respectively. We choose scenes for Replica to be in aligned with a previous researches. For the ScanNet, scenes were randomly selected based on least blur amount in RGB sequence. With the provided GT semantic segmentation annotation we perform 3D open-vocabulary benchmarking with the closest related works (Sec. \ref{ov_sem_seg_experiments}) to show advantage of BBQ 3D object-centric map scene representation. To examine the ability to answer complex queries that contain object relations we utilize ScanNet annotations from Sr3D+/Nr3D~\cite{achlioptas2020referit_3d} and ScanRefer~\cite{chen2020scanrefer} datasets. The Sr3D+ dataset contains template-based references to objects based on their spatial relations with other objects. The Nr3D and ScanRefer datasets contain various human-annotated natural language object references. For the template-based Sr3D+ dataset, we select all $661$ unique queries with triplets of $(target, relation, anchor)$ for the eight considered scenes from ScanNet. For the natural language Nr3D and ScanRefer datasets we select all queries from the eight considered scenes resulting in $699$ and $720$ queries respectively. Results are provided in Sec. \ref{scene_graph_results}. All the experiments were conducted on a single Nvidia V100 with 32 GB of vRAM except for LLM where we use Nvidia H100 with 80 GB of vRAM.

\subsection{3D open-vocabulary semantic segmentation}
\label{ov_sem_seg_experiments}

To perform 3D open-vocabulary segmentation we extract CLIP features for each object cropped view. Our ablation between OpenCLIP \cite{radford2021learning}, SigLip \cite{zhai2023sigmoid}, and Eva2 \cite{fang2023eva} shows that the last model performs better in terms of quality by a large margin. We utilize Eva2 \cite{yang2024fine} and call this variant BBQ-CLIP.

\textit{Evaluation protocol.}
For all methods, we query only classes that exist on each scene inside the phrase \textit{``an image of \textless class name\textgreater
"} and calculate mAcc, mIoU, and frequency-weighted mIoU.

\textit{Results.}
As shown in Tab. \ref{tab:3Dsemseg}, our BBQ-CLIP approach shows best results among other zero-shot algorithms (ConceptFusion\cite{conceptfusion}, ConceptGraph\cite{gu2023conceptgraphs}, OpenMask3D\cite{takmaz2023openmask3d}) in the 3D open-vocabulary segmentation and compeat with OpenFusion\cite{yamazaki2023open}. We position OpenFusion in a privileged group because the method exploits SEEM \cite{zou2024segment}, which was trained on supervised segmentation tasks in evaluation domain.

\begin{table}[t]
  \tiny
  \caption{Graph edges ablation study on Nr3D dataset (GT objects).}
  \label{tab:object_retrieval_gt}
  \centering
  \begin{tabularx}{0.45\textwidth}{p{1.2cm}p{1.3cm}p{1cm}p{1.25cm}p{1.25cm}}
    \toprule
    LLM & Edge & Recall@1 (Overall) & Recall@1 (View Independent) & Recall@1 (View Dependent) \\
    \midrule
    Llama3-8B & - & 36.1 & 36.1 & 36.0 \\
    Llama3-8B & Metric & \underline{43.8} & \underline{43.0} & 46.3 \\
    Llama3-8B & Semantic & 41.4 & 37.8 & \textbf{53.0} \\
    Llama3-8B & Metric+Semantic & \textbf{45.5} & \textbf{43.3} & \underline{52.4} \\
    \midrule
    GPT-4o & - & 61.8 & 67.8 & 43.7 \\
    GPT-4o & Metric & \textbf{68.6} & \textbf{73.9} & 52.4 \\
    GPT-4o & Semantic & 50.5 & 49.2 & \underline{54.9} \\
    \textbf{GPT-4o} & Metric+Semantic & \underline{68.4} & \underline{70.3} & \textbf{62.1} \\
  \bottomrule
  \end{tabularx}
\end{table}

\begin{table}[t]
  \tiny
  \caption{
    Grounding accuracy on Sr3D+/Nr3D dataset.
  }
  \label{tab:object_retrieval_sr3d_nr3d}
  \centering
  \begin{tabularx}{0.45\textwidth}{p{0.85cm}p{0.2cm}p{0.2cm}p{0.25cm}p{0.25cm}p{0.25cm}p{0.25cm}p{0.25cm}p{0.25cm}p{0.25cm}p{0.25cm}}
    \toprule 
      \multicolumn{1}{c}{} &
      \multicolumn{10}{c}{Sr3D+} \\
    \midrule
      \multicolumn{1}{c}{} &
      \multicolumn{2}{c}{Overall} &
      \multicolumn{2}{c}{Easy} &
      \multicolumn{2}{c}{Hard} &
      \multicolumn{2}{c}{View Dep.} &
      \multicolumn{2}{c}{View Indep.} \\
    Methods & A@0.1 & A@0.25 & A@0.1 & A@0.25 & A@0.1 & A@0.25
    & A@0.1 & A@0.25 & A@0.1 & A@0.25 \\
    \midrule
    OpenFusion  & 12.6 & 2.4 & 14.0 & 2.4 & 1.3 & 1.3 & 3.8 & 2.5 & 13.7 & 2.4 \\
    BBQ-CLIP  & 14.4 & 8.8 & 15.4 & 9.0 & 6.7 & 6.7 & 11.4 & 5.1 & 14.4 & 8.8 \\
    ConceptGraphs  & 13.3 & 6.2 & 13.0 & 6.8 & 16.0 & 1.3 & 15.2 & 5.1 & 13.1 & 6.4 \\
    \textbf{BBQ}  & \textbf{34.2} & \textbf{22.7} & \textbf{34.3} & \textbf{22.7} & \textbf{33.3} & \textbf{22.7} & \textbf{32.9} & \textbf{20.3} & \textbf{34.4} & \textbf{23.0} \\
    \midrule
      \multicolumn{1}{c}{} &
      \multicolumn{10}{c}{Nr3D} \\
    \midrule
      \multicolumn{1}{c}{} &
      \multicolumn{2}{c}{Overall} &
      \multicolumn{2}{c}{Easy} &
      \multicolumn{2}{c}{Hard} &
      \multicolumn{2}{c}{View Dep.} &
      \multicolumn{2}{c}{View Indep.} \\
    Methods & A@0.1 & A@0.25 & A@0.1 & A@0.25 & A@0.1 & A@0.25
    & A@0.1 & A@0.25 & A@0.1 & A@0.25 \\
    \midrule
    OpenFusion  & 10.7 & 1.4 & 12.9 & 1.4 & 5.1 & 1.5 & 8.5 & 0.0 & 11.4 & 1.9 \\
    BBQ-CLIP  & 15.3 & 9.4 & 18.1 & 11.0 & 8.1 & 5.6 & 8.1 & 6.1 & 17.2 & 10.5 \\
    ConceptGraphs  & 16.0 & 7.2 & 18.7 & 9.2 & 9.1 & 2.0 &  12.7 & 4.2 & 17.0 & 8.1 \\
    \textbf{BBQ}  & \textbf{28.3} & \textbf{19.0} & \textbf{30.5} & \textbf{21.3} & \textbf{22.8} & \textbf{13.2} & \textbf{23.6} & \textbf{18.2} & \textbf{29.8} & \textbf{19.3} \\
  \bottomrule
  \end{tabularx}
\end{table}

\begin{table}[t]
  \scriptsize
  \caption{Grounding accuracy on ScanRefer dataset.}
  \label{tab:object_retrieval_scanrefer}
  \centering
  \begin{tabularx}{0.45\textwidth}{XXX}
    \toprule 
    Methods & A@0.25 & A@0.5 \\
    \midrule
    LERF & 4.4 & 0.3 \\
    OpenScene  & 13.0 & 5.1\\
    LLM-Grounder & 17.1 & 5.3 \\
    \textbf{BBQ} & \textbf{19.4} & \textbf{11.6} \\
  \bottomrule
  \end{tabularx}
\end{table}

\subsection{Scene graphs generation}
\label{scene_graph_results}
\textit{Evaluation protocol.}
During ablation experiments, we assess various methods for constructing a graph from ground-truth point clouds of objects. We evaluate object grounding quality using the $Recall@1$ ~\cite{gu2023conceptgraphs}. For experiments involving scene reconstruction, we use such metrics as $Acc@0.1$~\cite{werby2024hierarchical}, $Acc@0.25$ and $Acc@0.5$~\cite{chen2020scanrefer}. A prediction is considered true positive if the Intersection over Union (IoU) between the selected object bounding box and the ground truth bounding box exceeds $0.1$, $0.25$ and $0.5$, respectively.

\textit{Results.}
The Nr3D dataset annotates referring expressions as containing view-dependent and view-independent spatial relations. We conduct experiments on combining metric and semantic spatial edges to explore how the edge type affects the quality of object grounding for different types of queries. We use LLAMA3-8B~\cite{llama3modelcard} and GPT4-o~\cite{achiam2023gpt} (\textit{gpt4o-2024-08-06}) in this study. For both models, adding semantic spatial edges improves the quality of object grounding compared to absence of edges and metric edges when the query contains view-dependent relations. Using a combination of metric and semantic spatial edges improves the quality for view-independent queries compared to using only semantic spatial edges for both models, so we use this combination in further experiments.

Our approach surpasses existing open-vocabulary methods for 3D object grounding, notably outperforming ConceptGraphs~\cite{gu2023conceptgraphs}, which also integrates an LLM, with scene text descriptions on Sr3D+ (Table~\ref{tab:object_retrieval_sr3d_nr3d}) and Nr3D (Table~\ref{tab:object_retrieval_sr3d_nr3d} and Table~\ref{tab:object_retrieval_nr3d_color}). 
In these experiments, we use the GPT4-o~\cite{achiam2023gpt} (\textit{gpt4o-2024-08-06}) for all methods.
It is worth noting that our method outperforms baseline approaches for various types of queries, including different types of spatial relations (view dependent and view-independent), and mentions of color, shape. BBQ can process queries that do not mention target class explicitly, i.e. the object described by its function (Table~\ref{tab:object_retrieval_nr3d_color}). 
Additionally, we compare our method with the baseline approaches LERF \cite{kerr2023lerf}, OpenScene \cite{peng2023openscene}, LLM-Grounder\cite{yang2024llm} on the same ScanRefer subset used in the experiments in LLM-Grounder\cite{yang2024llm}. This subset consists of $14$ ScanNet scenes and $998$ referring expressions. Table~\ref{tab:object_retrieval_scanrefer} demonstrates the effectiveness of our approach compared to these baseline approaches.

Moreover, we demonstrate that leveraging an object-centric textual description of a scene graph, interfaced via an LLM, significantly outperforms methods utilizing CLIP features for object grounding, such as OpenFusion~\cite{yamazaki2023open} and our BBQ-CLIP. However, in some practical cases such approach can be usefull, especially if there are limited resources to call LLM.

\begin{table*}[t]
  \tiny
  \caption{
    Grounding accuracy on Nr3D dataset.
  }
  \label{tab:object_retrieval_nr3d_color}
  \centering
  \begin{tabularx}{1.0\textwidth}{|p{1.2cm}|p{0.45cm}p{0.45cm}|p{0.45cm}p{0.45cm}|p{0.45cm}p{0.45cm}|p{0.45cm}p{0.45cm}|p{0.45cm}p{0.45cm}|p{0.45cm}p{0.45cm}|p{0.45cm}p{0.45cm}|p{0.45cm}p{0.45cm}|p{0.45cm}p{0.45cm}|}
    \toprule 
      \multicolumn{1}{|c|}{} &
      \multicolumn{2}{c|}{Overall} &
      \multicolumn{2}{c|}{w Spatial Lang.} &
      \multicolumn{2}{c|}{w/o Spatial Lang.} &
      \multicolumn{2}{c|}{w Color Lang.} &
      \multicolumn{2}{c|}{w/o Color Lang.} &
      \multicolumn{2}{c|}{w Shape Lang.} &
      \multicolumn{2}{c|}{w/o Shape Lang.} &
      \multicolumn{2}{c|}{w Target Mention} &
      \multicolumn{2}{c|}{w/o Target Mention}\\
    Methods & A@0.1 & A@0.25 & A@0.1 & A@0.25 & A@0.1 & A@0.25
    & A@0.1 & A@0.25 & A@0.1 & A@0.25 & A@0.1 & A@0.25
    & A@0.1 & A@0.25 & A@0.1 & A@0.25 & A@0.1 & A@0.25\\
    \midrule
    OpenFusion  & 10.7 & 1.4 & 8.9 & 1.1 & 22.3 & 3.2 & 11.8 & 0.8 & 10.5 & 1.6 & 9.8 & 1.0 & 10.9 & 1.5 & 11.3 & 1.6 & 4.9 & 0.0 \\
    BBQ-CLIP  & 15.3 & 9.4 & 14.7 & 8.8 & 19.1 & 13.8 & 24.4 & 16.8 & 13.4 & 7.9 & 12.7 & 6.9 & 15.7 & 9.9 & 16.3 & 10.0 & 4.9 & 3.3 \\
    ConceptGraphs  & 16.0 & 7.2 & 15.0 & 6.6 & 22.3 & 10.6 & 17.6 & 5.9 & 15.7 & 7.4 & 10.8 & 4.9 & 16.9 & 7.5 & 16.9 & 7.5 & 6.6 & 3.3 \\
    \textbf{BBQ}  & \textbf{28.3} & \textbf{19.0} & \textbf{28.1} & \textbf{19.2} & \textbf{29.8} & \textbf{18.1} & \textbf{25.2} & \textbf{17.6} & \textbf{29.0} & \textbf{19.3} & \textbf{34.3} & \textbf{23.5} & \textbf{27.3} & \textbf{18.3} & \textbf{29.6} & \textbf{19.7} & \textbf{14.8} & \textbf{11.5} \\
  \bottomrule
  \end{tabularx}
\end{table*}

\begin{figure}[t]
   \centering
   \includegraphics[width=0.5\textwidth]{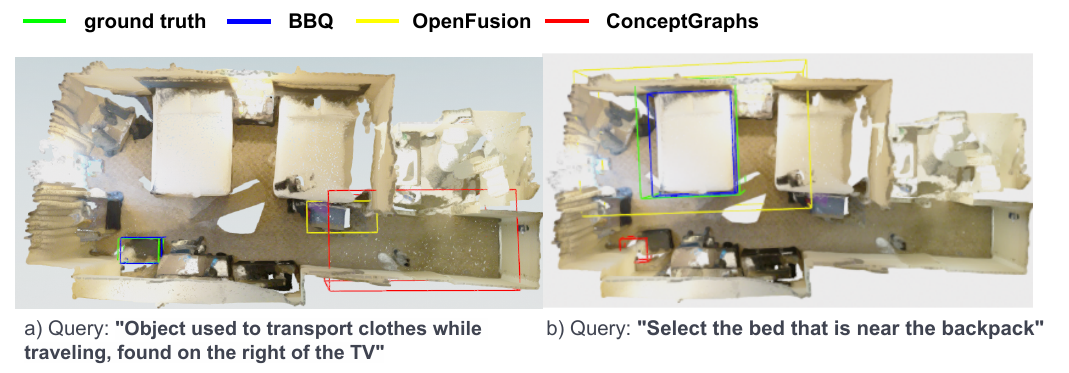}
   \vspace{-5mm}
   \caption{\label{qual_results} Qualitative examples of 3D referred object grounding on the Sr3D+/Nr3D datasets.}
\end{figure}

\subsection{Real-World Experiments}
\label{real_robot}

\begin{figure}
\begin{minipage}[t]{0.45\columnwidth}
  \centering
  \includegraphics[width=\textwidth]{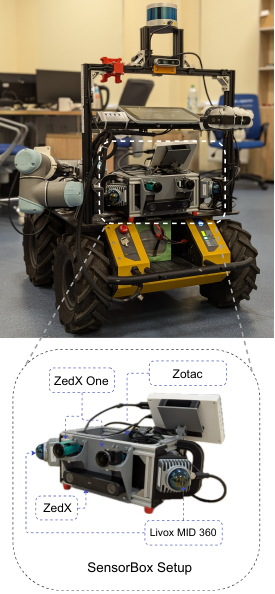}
  \vspace{-5mm}
  \caption{\label{sensorbox} Husky mobile robot with SensorBox used for real-world experiments}
\end{minipage}\hfill % maximize horizontal separation
\begin{minipage}[t]{0.5\columnwidth}
  \centering
  \includegraphics[width=\textwidth]{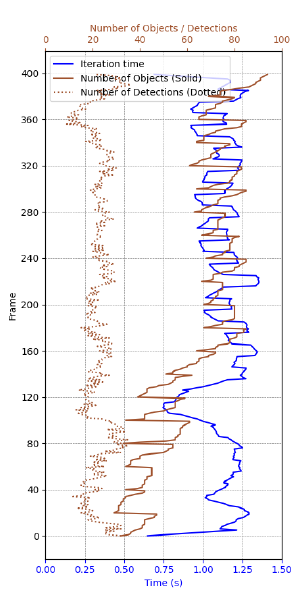}
  \vspace{-5mm}
  \caption{\label{zotac_time} 3D mapping process speed on the Zotac on-board computer with Replica room0 RGB-D sequence}
\end{minipage}
\end{figure}

BBQ was tested using custom SensorBox hardware (Fig.~\ref{sensorbox}). The SensorBox contains the following sensors: one stereo RGB Zed X camera with integrated IMUs, two LiDARs and two fisheye RGB cameras. The SensorBox is constructed on a rigid metal frame with a built-in LiPo battery and an on-board PC Zotac (11th Gen i7-11800H @2.3 GHz, Nvidia GeForce RTX3080 16GB Mobile, 62.5 GB RAM), thereby ensuring the real-time data acquisition.

First, we measured BBQ mapping performance on our hardware with publicly available data. Because performance relies on the number of objects, we chose room0 RGB-D sequence with stride 10 from the Replica dataset for benchmarking. As demonstrated in Fig.~\ref{zotac_time}, BBQ works on the on-board Zotac PC with considerable mean time of 1.18 s/it and an overall time of 7 min 52 seconds for scene. It is almost $\times3$ faster than ConceptGraphs \cite{gu2023conceptgraphs}.

Second, we conducted experiments on real-world data in our facilities. We equipped a mobile platform, Husky, with the SensorBox (Fig.~\ref{sensorbox}) and performed whole pipeline testing. We reconstructed pose with vSLAM \cite{campos2021orb} and generate depth images with the neural network \cite{li2022practical}. The 3D mapping process was launched onboard, but vLLM and LLM were deployed on server with an Nvidia V100 32Gb vRAM with API. The success of the performed experiments demonstrated BBQ's ability to be deployed in a real-life scenario. We have added examples of our algorithm running on real data to the supplementary materials.

\section{Limitations}
\label{limitations}

It should be noted that BBQ works under the assumption of a static environment equivalent to a standard indoor room. To handle dynamic, existing tracking methods~\cite{karaev2024cotracker,stanojevic2024boosttrack++,tumanyan2024dino} should be applied. For large and complex form locations, 3D scene graph in current implementation may not reflect semantic spatial relations correctly. 

Also, conducted experiments highlight that our 3D object-centric map construction method is limited in its ability to successfully distinguish tiny objects in the image. It is a general problem of visual foundation models, which can be partially solved with an integration of target-specific detectors~\cite{muzammul2025comprehensive} or with more scene exploration where the camera is placed closer to objects of interest to successfully map such instances.

\section{Conclusion}

With BBQ, we advance the limits of 3D scene perception by integrating language models with general world knowledge and our scene-specific graph representation. The successful outcomes of our experiments on the complex Nr3D, Sr3D, and ScanRefer datasets, as well as on real-life data, demonstrate the effectiveness of our approach with metric and semantic scene edges, opening new avenues for a more comprehensive and flexible understanding and interaction with 3D scenes. We hope that our resource-efficient code implementation will facilitate BBQ applications in real-world robotics projects that bridge the communication gap between humans and intelligent autonomous agents.

\section*{ACKNOWLEDGMENT}

We thank team of Sberbank Robotics Center for supporting us in conducting real-world robotic experiments.

% %%%%%%%%%%%%%%%%%%%%%%%%%%%%%%%%%%%%%%%%%%%%%%%%%%%%%%%%%%%%%%%%%%%%%%%%%%%%%%%%

% References are important to the reader; therefore, each citation must be complete and correct. If at all possible, references should be commonly available publications.
\newpage

\bibliographystyle{IEEEtran}
\bibliography{IEEEabrv,bib}

\end{document}